# Accuracy Improvement for Stiffness Modeling of Parallel Manipulators

A. Pashkevich[1,2], A. Klimchik[1,2], D. Chablat[1], Ph. Wenger[1]
[1]Institut de Recherches en Communications et Cybernetique de Nantes, 1 rue de la No, 44321 Nantes, France
[2]Ecole des Mines de Nantes, 4 rue Alfred-Kastler, Nantes 44307, France

**Abstract**
The paper focuses on the accuracy improvement of stiffness models for parallel manipulators, which are employed in high-speed precision machining. It is based on the integrated methodology that combines analytical and numerical techniques and deals with multidimensional lumped-parameter models of the links. The latter replace the link flexibility by localized 6-dof virtual springs describing both translational/rotational compliance and the coupling between them. There is presented detailed accuracy analysis of the stiffness identification procedures employed in the commercial CAD systems (including statistical analysis of round-off errors, evaluating the confidence intervals for stiffness matrices). The efficiency of the developed technique is confirmed by application examples, which deal with stiffness analysis of translational parallel manipulators

## 1 INTRODUCTION

Currently, parallel manipulators have become more and more popular for a variety of technological processes, including high-speed precision machining [1] [2]. This growing attention is inspired by their essential advantages over serial manipulators, which have already reached the dynamic performance limits. In contrast, parallel manipulators are claimed to offer better accuracy, lower mass/inertia properties, and higher structural rigidity (i.e. stiffness-to-mass ratio) [3].

These features are induced by their specific kinematic structure, which resists the error accumulation in kinematic chains and allows convenient actuators location close to the manipulator base. This makes them attractive for innovative robotic systems, but practical utilization of the potential benefits requires development of efficient stiffness analysis techniques, which satisfy the computational speed and accuracy requirements of relevant design procedures.

Generally, the stiffness analysis evaluates the effect of the applied external torques and forces on the compliant displacements of the end-effector. Numerically, this property is defined through the "stiffness matrix", which gives the relation between the translational/rotational displacement and the static forces/torques causing this transition [4]. Similar to other manipulator properties (kinematical, for instance), the stiffness essentially depends on the force/torque direction and on the manipulator configuration [5].

Several approaches exist for the computation of the stiffness matrix, such as the Finite Element Analysis (FEA), the matrix structural analysis (MSA), and the virtual joint method (VJM). The FEA method is proved to be the most accurate and reliable, since the links/joints are modeled with its true dimension and shape. Its accuracy is limited by the discretization step only. However, because of high computational expenses required for the repeated re-meshing, this method is usually is not very popular in industrial robotics.

The MSA method incorporates the main ideas of the FEA but operates with rather large flexible elements (beams, arcs, cables, etc.). This obviously yields reduction of the computational expenses and, in some cases, allows even obtaining an analytical stiffness matrix. This method gives a reasonable trade-off between the accuracy and computational time, provided that link approximation by the beam elements is realistic.

Finally, the VJM method, which is also referred to as the "lumped modeling", is based on the expansion of the traditional rigid model by adding virtual joints, which describe the elastic deformations of the manipulator components (links, joints and actuators). This approach originates from the work of Gosselin [6], who evaluated parallel manipulator stiffness taking into account only the actuators compliance. At present, there are a number of variations and simplifications of the VJM method, which differ in modeling assumptions and numerical techniques.

Recent modification of this method proposed by the authors [7] allows to extend it to the over-constrained manipulator and to apply it at any workspace point, including the singular ones. The method is based on a multidimensional lumped-parameter model that replaces the link flexibility by localized 6-dof virtual springs that describe both the linear/rotational deflections and the coupling between them. The spring stiffness parameters are evaluated using FEA modelling to take into account real shape of the manipulator components. This gives almost the same accuracy as FEA but with essentially lower computational effort because it eliminates re-meshing through the workspace.

This paper focuses on the accuracy improvement of the VJM method via enhancement of the identification algorithms incorporated in FEA-based stiffness evaluation. This leads to adequate modeling of the manipulator components and allows essentially reduce the modeling errors for the whole mechanism. In contrast to previous works, the proposed technique operates with the deflection field composed of the set of the nodes and allows evaluate statistical significance of the estimated elements of the stiffness matrix.

The reminder of the paper is organized as follows. In the Section 2, it is defined the set of problems which are considered in the paper. Section 3 presents a new method for the deflections identification from the field of points. Section 4 illustrates the proposed method via a numerical example. Section 5 presents the stiffness identifications of the Orthoglide manipulator using the new method. And, finally, Section 6 summarizes the main contributions of the paper.

## 2 PROBLEM STATEMENT

The stiffness model describes the resistance of an elastic body or mechanism to deformations caused by an external force or torque. For relatively small deformations, this property is defined through the "stiffness matrix" **K**, which defines the linear relation

$$\begin{bmatrix} \mathbf{F} \\ \mathbf{M} \end{bmatrix} = \mathbf{K} \cdot \begin{bmatrix} \mathbf{t} \\ \boldsymbol{\varphi} \end{bmatrix} \qquad (1)$$

between the three-dimensional translational/rotational displacements $\mathbf{t} = (t_x, t_y, t_z)^T$ ; $\boldsymbol{\varphi} = (\varphi_x, \varphi_y, \varphi_z)^T$ and the static forces/torques $\mathbf{F} = (F_x, F_y, F_z)$ , $\mathbf{M} = (M_x, M_y, M_z)$ causing this transition. As known from mechanics, $K$ is a 6×6 symmetrical semi-definite non-negative matrix, which may include non-diagonal elements to represent the coupling between the translations and rotations [7]. The inverse of **K** is usually called the "compliance matrix" and is denoted as *k*.

For robotic manipulators, the matrix *K* can be computed semi-analytically provided that the stiffness matrices of all separate components (links, actuators, etc.) are known with desired precision [7]. However, explicit expressions for the link stiffness matrices can be obtained in simple cases only (truss, beam, etc.). For more sophisticated shapes that are commonly used in robotics, the stiffness matrix is usually estimated via the shape approximation, using relatively small set of primitives [8]. However, as follows from our previous study [7], accuracy of this approach is rather low (errors from 30% to 50%). Hence, in general case, it is prudent to apply to each link the FEA-based techniques, which hypothetically produce rather accurate result.

Using the FEA, the stiffness matrix **K** (or its inverse **k**) is evaluated from several numerical experiments, each of which produces the vectors of linear and angular deflections (**t**, **φ**) corresponding to the applied force and torque (**F**, **M**). Then, the desired matrix is computed from the linear system

$$\mathbf{k} = \begin{bmatrix} \mathbf{F}_1 & \cdots & \mathbf{F}_m \\ \mathbf{M}_1 & \cdots & \mathbf{M}_m \end{bmatrix}^{-1} \cdot \begin{bmatrix} \mathbf{t}_1 & \cdots & \mathbf{t}_m \\ \boldsymbol{\varphi}_1 & \cdots & \boldsymbol{\varphi}_m \end{bmatrix} \qquad (2)$$

where *m* is the number of experiments ($m \geq 6$) and the matrix inverse is replaced by the pseudoinverse in the case of $m > 6$. It is obvious that numerically attractive is the case of $m = 6$ with special arrangement of the forces and torques

$\mathbf{F}_1 = [F_x, 0, 0]^T$ ; $\mathbf{M}_1 = [0, 0, 0]^T$ ;
$\mathbf{F}_2 = [0, F_y, 0]^T$ ; $\mathbf{M}_2 = [0, 0, 0]^T$ ;
… …
$\mathbf{F}_6 = [0, 0, 0]^T$ ; $\mathbf{M}_6 = [0, 0, M_z]^T$

corresponding to the diagonal structure of the matrix to be inverted. In this case, each FEA-experiment produces exactly one column of the compliance matrix

$$\mathbf{k} = \begin{bmatrix} \mathbf{t}_1/F_x & \cdots & \mathbf{t}_m/M_z \\ \boldsymbol{\varphi}_1/F_x & \cdots & \boldsymbol{\varphi}_m/M_z \end{bmatrix} \qquad (3)$$

and the values $(\mathbf{t}_i, \boldsymbol{\varphi}_i)$ may be clearly physically interpreted. On the other hand, by increasing the number of experiments ($m > 6$) it is possible reduce the estimation error, which is in the focus of this paper.

It is obvious that the main source of estimation errors is related to the FEA-modeling that highly depends on the size and type of the finite elements, meshing options, incorporated numerical algorithms, computer word length and round-off principle. Hypothetically, the accuracy can be essentially improved by reducing the mesh size and increasing the number of digits in presentation of all variables. But there are some evident technical constraints that do not allow ignoring the FEA limitations.

Another type of errors arises from numerical differentiation incorporated in the considered technique. Strictly speaking, the linear relation (1) is valid for rather small deflections that may be undetectable against the FEA-modeling defects. On the other side, large deflections may be out of the elasticity range. Hence, it is prudent to find compromise for the applied forces/torques taking into account both factors.

In order to increase accuracy; it is also worth to improve the deflection estimation technique. Traditionally, the values (**t**, **φ**) are computed from spatial location of a single finite element enclosing the reference point (RP). In contrast to this approach, it is proposed to evaluate (**t**, **φ**) from the *displacement field* describing transitions of rather large number of nodes located in the neighborhood of RP. It is reasonable to assume that such modification will yield positive result, since the FEA-modeling errors are usually differ from node to node, exposing almost quasi-stochastic nature.

To formulate this problem strictly, let us denote the displacement field by a set of vector couples $\{\mathbf{p}_i, \Delta\mathbf{p}_i \mid i = \overline{1,n}\}$ where the first component $\mathbf{p}_i$ define the node initial location (before applying the force/torque), $\Delta\mathbf{p}_i$ refers to the node displacement due to the applied force/torque, and *n* is the number of considered nodes. Then, assuming that all the nodes are close enough to the reference point, this set can be approximated by a "*rigid transformation*"

$$\mathbf{p}_i + \Delta\mathbf{p}_i = \mathbf{R}(\boldsymbol{\varphi}) \cdot \mathbf{p}_i + \mathbf{t} , i = \overline{1,n} \qquad (4)$$

that includes as the parameters the linear displacement **t** and the orthogonal 3×3 matrix *R* that depends on the rotational displacement **φ**. Then, the problem of the deflection estimation can be presented as the best fit of the considered vector field by equation (4) with respect to six scalar variables incorporated in **t**, **R**.

In practice, the FEA-modeling output provides the deflection vector fields for all nodes referring to all components of the mechanism. So, it is required to select relevant subset corresponding to the neighborhood of the reference point $\mathbf{p}_0$. Besides, the node locations $\mathbf{p}_i$ must be expressed relative to this point, i.e. the origin of the coordinate system must be shifted to $\mathbf{p}_0$. The latter is specified by the physical meaning of the deflections in the stiffness analysis.

Thus, the primary problem to be solved in this paper is the development of efficient numerical technique for estimation of the deflections from the vector field. Besides, there are a number of subsidiary problems to consider, including: (i) defining reasonable size of finite elements, selecting meshing options and the assigning prudent force/torque amplitudes for the FEA modeling; (ii) developing simple rules for constructing the deflection field ('rule of thumb' for RP-neighborhood); (iii) elaboration of filtering techniques allowing to eliminate outliers in the FEA output; (iv) significance testing for the identified parameters and estimation their confidence intervals.

## 3 DEFLECTIONS IDENTIFICATION FROM FEA FIELD

To estimate the desired deflections ($\mathbf{t}, \boldsymbol{\varphi}$), let us apply the least square technique that leads to minimization of the sum of squared residuals

$$f = \sum_{i=1}^{n} \| \mathbf{p}_i + \Delta\mathbf{p}_i - \mathbf{R}(\boldsymbol{\varphi})\mathbf{p}_i - \mathbf{t} \|^2 \to \min_{\mathbf{R},\mathbf{t}} \quad (5)$$

with respect to the vector $\mathbf{t}$ and the orthogonal matrix $\mathbf{R}$ representing the rotational deflections $\boldsymbol{\varphi}$. The specificity of this problem (that does not allow direct application of the standard methods) are the orthogonally constraint $\mathbf{R}^T\mathbf{R} = \mathbf{I}$ and non-trivial relation between elements of the matrix $\mathbf{R}$ and the vector $\boldsymbol{\varphi}$. The following subsections presents two methods for computing $\mathbf{t}, \boldsymbol{\varphi}$, as well as their comparison study.

### 3.1 SVD-based method

This technique was elaborated in our previous paper [7] and relies on some results from matrix algebra referred to the orthogonal "Procrustes problem" [9]. The estimation procedure is decomposed in two steps, which sequentially produce the rotation matrix $\mathbf{R}$ and the translation vector $\mathbf{t}$. Then, the desired vector of rotation angles $\boldsymbol{\varphi}$ is extracted from $\mathbf{R}$ using linearization. Let us briefly present the mathematical background paying primary attention to the accuracy issues.

Here, the desired solution is obtained by minimization of the function (5) subject to $\mathbf{R}^T\mathbf{R} = \mathbf{I}$. First, for the non-constrained variable $\mathbf{t}$, straightforward differentiation and equating to zero gives an expression

$$\mathbf{t} = \frac{1}{n}\left(\sum_{i=1}^{n} \Delta\mathbf{p}_i - (\mathbf{R} - \mathbf{I})\sum_{i=1}^{n} \mathbf{p}_i\right) \quad (6)$$

Then, after relevant substitution and denoting

$$\hat{\mathbf{p}}_i = \mathbf{p}_i - \frac{1}{n}\sum_{i=1}^{n}\mathbf{p}_i; \quad \hat{\mathbf{g}}_i = \mathbf{p}_i + \Delta\mathbf{p}_i - \frac{1}{n}\sum_{i=1}^{n}(\mathbf{p}_i + \Delta\mathbf{p}_i), \quad (7)$$

the original optimization problem is reduced to the orthogonal Procrustes formulation]

$$f = \sum_{i=1}^{n} \| \hat{\mathbf{g}}_i - \mathbf{R}\hat{\mathbf{p}}_i \|^2 \to \min_{\mathbf{R}}. \quad (8)$$

The latter yields the solution [9]

$$\mathbf{R} = \mathbf{V}\mathbf{U}^T \quad (9)$$

that is expressed via the singular value decomposition (SVD) of the matrix

$$\sum_{i=1}^{n} \hat{\mathbf{p}}_i \hat{\mathbf{g}}_i^T = \mathbf{U}\boldsymbol{\Sigma}\mathbf{V}^T, \quad (10)$$

which requires some computational efforts. Hence, the above expressions (6), (9) allow to solve the optimization problem (5) in terms of variables $\mathbf{R}$ and $\mathbf{t}$.

Further, to evaluate the vector $\boldsymbol{\varphi}$, the orthogonal matrix $\mathbf{R}$ must be decomposed into product of elementary rotations around the Cartesian axes x, y, z. It is obvious that, in general case, this decomposition is not unique and depends on the rotation order. However, for small $\varphi$ (that is implicitly assumed for FEA-experiments) this matrix may be uniquely presented in differential form as

$$\mathbf{R} \cong \begin{bmatrix} 1 & -\varphi_z & \varphi_y \\ \varphi_z & 1 & -\varphi_x \\ -\varphi_y & \varphi_x & 1 \end{bmatrix} \quad (11)$$

| Method | $\varphi_x$ | $\varphi_y$ | $\varphi_z$ |
|---|---|---|---|
| SVD+ | $r_{32}$ | $r_{13}$ | $r_{21}$ |
| SVD− | $-r_{23}$ | $-r_{31}$ | $-r_{12}$ |
| SVD± | $(r_{32}-r_{23})/2$ | $(r_{13}-r_{31})/2$ | $(r_{21}-r_{12})/2$ |

Table 1: Evaluation of the rotation angles from matrix $\mathbf{R}$

Using this expression, the desired parameters $\varphi_x, \varphi_y, \varphi_z$ may be extracted from $\mathbf{R} = [r_{ij}]$ in several ways (Table 1), which are formally equivalent but do not necessarily posses similar robustness with respect to round-off errors. Relevant comparison study is presented in subsection 3.3.

### 3.2 LIN-based method.

To reduce the computational efforts and to avoid the SVD, let us introduce linearization of the rotational matrix $\mathbf{R}$ at the early stage, using explicit parameterization given by expression (11). This allows to rewrite equation of the 'rigid transformation' (4) in the form

$$\Delta\mathbf{p}_i = \mathbf{p}_i \times \boldsymbol{\varphi} + \mathbf{t}; \quad i = \overline{1,n} \quad (12)$$

that can be further transformed into a linear system of the following form

$$\begin{bmatrix} \mathbf{I} & \mathbf{P}_i \end{bmatrix} \begin{bmatrix} \mathbf{t} \\ \boldsymbol{\varphi} \end{bmatrix} = \Delta\mathbf{p}_i; \quad i = \overline{1,n} \quad (13)$$

where $\mathbf{P}_i$ is a skew-symmetric matrix corresponding to the vector $\mathbf{p}_i$:

$$\mathbf{P}_i = \begin{bmatrix} 0 & p_{zi} & -p_{yi} \\ -p_{zi} & 0 & p_{xi} \\ p_{yi} & -p_{xi} & 0 \end{bmatrix} \quad (14)$$

Then, applying the standard least-square technique with the objective

$$f = \sum_{i=1}^{n} \| \Delta\mathbf{p}_i - \mathbf{P}_i\boldsymbol{\varphi} - \mathbf{t} \|^2 \to \min_{\boldsymbol{\varphi},\mathbf{t}} \quad (15)$$

one can get the solution

$$\begin{bmatrix} \mathbf{t} \\ \boldsymbol{\varphi} \end{bmatrix} = \begin{bmatrix} n\mathbf{I} & \sum_{i=1}^{n}\mathbf{P}_i \\ \sum_{i=1}^{n}\mathbf{P}_i^T & \sum_{i=1}^{n}\mathbf{P}_i^T\mathbf{P}_i \end{bmatrix}^{-1} \begin{bmatrix} \sum_{i=1}^{n}\Delta\mathbf{p}_i \\ \sum_{i=1}^{n}\mathbf{P}_i^T\Delta\mathbf{p}_i \end{bmatrix} \quad (16)$$

that employs the 6×6 matrix inversion. This solution can be simplified by shifting the origin of the coordinate system to the point $\mathbf{p}_c = n^{-1}\sum_{i=1}^{n}\mathbf{p}_i$ leading to expression

$$\begin{bmatrix} \mathbf{t} \\ \boldsymbol{\varphi} \end{bmatrix} = \begin{bmatrix} n^{-1}\mathbf{I} & \mathbf{0} \\ \mathbf{0} & \left(\sum_{i=1}^{n}\hat{\mathbf{P}}_i^T\hat{\mathbf{P}}_i\right)^{-1} \end{bmatrix} \cdot \begin{bmatrix} \sum_{i=1}^{n}\Delta\mathbf{p}_i \\ \sum_{i=1}^{n}\hat{\mathbf{P}}_i^T\Delta\mathbf{p}_i \end{bmatrix} \quad (17)$$

that requires inversion of the matrix of size 3×3. Here, following the adopted notation, $\hat{\mathbf{P}}_i$ is a skew-symmetric matrix corresponding to the vector $\hat{\mathbf{p}}_i = \mathbf{p}_i - \mathbf{p}_c$. Let us also consider several cases that are useful for practical applications.

**Case 1: Symmetrical field.** If the field is symmetrical with respect to its centre $\mathbf{p}_c$, the solution (17) can be presented in compact analytical form as

$$\mathbf{t} = n^{-1}\sum_{i=1}^{n}\Delta\mathbf{p}_i; \quad \boldsymbol{\varphi} = \mathbf{D}^{-1}\sum_{i=1}^{n}\hat{\mathbf{P}}_i^T\Delta\mathbf{p}_i$$

where the matrix

$$\mathbf{D} = diag\left[\sum_{i=1}^{n}(\hat{p}_{yi}^2 + \hat{p}_{zi}^2) \quad \sum_{i=1}^{n}(\hat{p}_{xi}^2 + \hat{p}_{zi}^2) \quad \sum_{i=1}^{n}(\hat{p}_{xi}^2 + \hat{p}_{yi}^2)\right]$$

is diagonal and easily inverted.

**Case 2: Cubic field.** If the field is symmetrical and, in addition, it is produced by uniform meshing of the cubic subspace $a \times a \times a$, the matrix D is expressed as $\mathbf{D} = d \cdot \mathbf{I}$ where $d = a^2 n(\sqrt[3]{n}-1)/6(\sqrt[3]{n}+1)$.

**Case 3: Planar square field.** For the filed produced by uniform meshing of the square $a \times a$ located perpendicular to the *x*-axis, the expression for the matrix $\mathbf{D} = diag[d \quad d/2 \quad d/2]$

The derived expressions are computationally attractive and allow simultaneous estimation both translational and rotational deflections from the FEA-produced field. Below they are evaluated for the precision and robustness.

### 3.3 Influence of linearization and round-offs

Both of the proposed algorithms involve numerous matrix multiplications that may accumulate the round-off errors. Besides, they employ the first-order approximation of the matrix **R** that may create another source of inaccuracy. Hence, it is prudent to obtain numerical assessments corresponding to a typical case study.

For these assessments, there were examined data sets corresponding to the cubic field of size $10 \times 10 \times 10$ mm$^3$ with the mesh step 1 mm (1331 points). The deflections were generated via the 'rigid transformation' (4) with the parameters $\mathbf{t} = (a,a,a)^T$ and $\boldsymbol{\varphi} = (b,b,b)^T$ presented in Tables 2, 3. All calculations were performed using the double precision floating-point arithmetic.

As follows from the analysis, the influence of the linearization and round-offs is negligible for the translation (the induced errors are less than $10^{-14}$ mm). In contrast, for the rotation, practically acceptable results may be achieved for rather small angular deflections that are less than 1.0° (the errors are up to 0.01°). The latter impose essential constraint on the amplitude of the forces/torques in the FEA-modelling that must ensure reasonable deflections.

| Method | Translation amplitude $a$ | | | |
|---|---|---|---|---|
| | 0.01 mm | 0.1 mm | 1.0 mm | 10 mm |
| SVD | $10^{-16}$ | $10^{-16}$ | $10^{-16}$ | $2 \cdot 10^{-14}$ |
| LIN | $10^{-16}$ | $10^{-16}$ | $10^{-16}$ | $3 \cdot 10^{-15}$ |

Table 2: Identification errors for the translation [mm]

| Method | Rotation amplitude $b$ | | | |
|---|---|---|---|---|
| | 0.01° | 0.1° | 1.0° | 5.0° |
| SVD+ | $2 \cdot 10^{-6}$ | $2 \cdot 10^{-4}$ | $2 \cdot 10^{-2}$ | 0.48 |
| SVD− | $2 \cdot 10^{-6}$ | $2 \cdot 10^{-4}$ | $2 \cdot 10^{-2}$ | 0.48 |
| SVD± | $1 \cdot 10^{-6}$ | $1 \cdot 10^{-4}$ | $1 \cdot 10^{-2}$ | 0.24 |
| LIN | $1 \cdot 10^{-6}$ | $1 \cdot 10^{-4}$ | $1 \cdot 10^{-2}$ | 0.24 |

Table 3: Identification errors for the rotation $\boldsymbol{\varphi}$ [deg]

Another conclusion concerns comparison of the SVD-based and LIN-based methods. It justifies advantages of the proposed LIN-based technique that provides the best robustness and lower computational complexity.

### 3.4. Influence of FEA-modeling errors

By its general principle, the FEA-modeling is an approximate method that produces some errors caused by the discretization. Beside, even for the perfect modeling, the deflections in the neighborhood of the reference point do not exactly obey the equation (4). Hence, it is reasonable to assume that the 'rigid transformation' (4) incorporates some random errors

$$\mathbf{p}_i + \Delta\mathbf{p}_i = \mathbf{R}(\varphi)\cdot\mathbf{p}_i + \mathbf{t} + \boldsymbol{\varepsilon}_i; \quad i = \overline{1,n} \quad (18)$$

that are supposed to be independent and identically distributed (i.i.d.) Gaussian random variables with zero-mean and standard deviation σ.

In the frame of this assumption, the expression for the deflections (17) can be rewritten as

$$\mathbf{t} = \mathbf{t}^o + n^{-1}\sum_{i=1}^{n}\boldsymbol{\varepsilon}_i; \quad \boldsymbol{\varphi} = \boldsymbol{\varphi}^o + \left(\sum_{i=1}^{n}\hat{\mathbf{P}}_i^T\hat{\mathbf{P}}_i\right)^{-1}\sum_{i=1}^{n}\hat{\mathbf{P}}_i^T\boldsymbol{\varepsilon}_i \quad (19)$$

where the superscript 'o' corresponds to the 'true' parameter value. This justifies usual properties of the adopted point-type estimator (17), which is obviously unbiased and consistent. Furthermore, the variance-covariance matrices for **t**, φ may be expressed as

$$\text{cov}[\mathbf{t}] = \frac{\sigma^2}{n}\mathbf{I}; \quad \text{cov}[\boldsymbol{\varphi}] = \sigma^2\left(\sum_{i=1}^{n}\hat{\mathbf{P}}_i^T\hat{\mathbf{P}}_i\right)^{-1} \quad (20)$$

allowing to evaluate the estimation accuracy using common confidence interval technique. As follows from (20), for the translational deflection **t** the identification accuracy is defined by the standard deviation $\sigma/\sqrt{n}$ and depends on the number of the points only. In contrast, for the rotational deflection, the spatial location of the points is a very important issue. In particular, for the cubic filed of the size $a \times a \times a$, the standard deviation of the rotation angles may be approximately expressed as $\sigma/a\sqrt{n/6}$.

Another practical question is related to *detecting zero elements* in the compliance matrix or, in other word, evaluating the statistical significance of the computed values compared to zero. Relevant statistical technique [10] operates with the p-values that may be easily converted in the form $k\sigma_a$, where k is usually from 3 to 5 and the subscript '$a$' refers to a particular component of the vectors **t**, φ.

To evaluate the standard deviation σ describing the random errors $\boldsymbol{\varepsilon}$, one may use the residual-based estimator obtained from the expression

$$\text{E}\left(\sum_{i=1}^{n}\|\mathbf{p}_i + \Delta\mathbf{p}_i - \mathbf{R}(\varphi)\cdot\mathbf{p}_i - \mathbf{t}\|^2\right) = (3n-6)\sigma^2. \quad (21)$$

The latter may be easily derived taking into account that, for each experiment, the deflection filed consist of *n* three-dimensional vectors that are approximated by the model containing 6 scalar parameters. Moreover, to increase accuracy, it is prudent to aggregate the squared residuals for all FEA-experiments and to make relevant estimation using the coefficient $(3n-6)m\sigma^2$, where *m* is the experiments number.

In addition, to increase accuracy and robustness, it is reasonable to eliminate outliers in the experimental data. They may appear in the FEA-field due to some anomalous causes, such as unsufficient meshing of some elements, violation of the boundary conditions in some arears of the mechanical joints, etc. The simplest and realible method that is adopted in this paper is based on the 'data filtering' with respect to the residials (i.e. eliminating cernain percertainge of the points with the highest residual values).

### 3.5 Simulation study: cubic field of deflections

To evaluate combined influence of various error sources, let us extend the simulation study from Sub-section 3.3 that focuses on the deflection identification from the cubic field of size $10\times10\times10$ mm$^3$ (1331 points, mesh step 1 mm). In particular, let us contaminate all deflections using the Gaussian noise with the s.t.d. $5\times10^{-5}$ mm that is a typical value discovered from the examined FEA data sets (Table 4). Similar to the pervious case, all calculations were performed using the double precision floating-point arithmetic (16 decimal digits).

Simulation results confirmed the main theoretical derivations of the previous subsections. The identification errors obey the normal distribution (Figure 1) but their dispersions should be evaluated taking into account some additional issues. Thus, the s.t.d. of the translational error is about $1.36\cdot10^{-6}$ mm and depends only on the FEA-induced component that is evaluated as $\sigma/\sqrt{n}\approx1.37\cdot10^{-6}$ mm. The influence of the linearization and round-offs is negligible here (this component less than $10^{-14}$ mm). Also, this type of the error does not depend on the translation amplitude.

In contrast, for the rotational deflections, there exists strong dependence on the amplitude (Figure 2). In particular, for the angular deflection 0.1°, the s.t.d. of the identification error is about $8.4\cdot10^{-5}$ deg, while the FEA-induced component is evaluated as $\sigma/a\sqrt{n/6}\approx1.8\cdot10^{-5}$ deg and the linearization component is about $8.8\cdot10^{-5}$ deg (see Table 2). Moreover, the simulation results allow define preferable values of the angular deflection that may be extracted from the FEA-data with the highest accuracy. They show that that the deflection angles should be in the range 0.01 …0.2° to ensure the identification accuracy about 0.2%

Thus, the proposed LIN-based algorithm (subsection 3.2) allows to identify the desired deflections (**t**, **φ**) with the required accuracy while possessing lower computational complexity than the known SVD-based technique.

## 4 ILLUSTRATIVE EXAMPLE

To demonstrate efficiency of the developed technique and to evaluate its applicability to real-life situations, let us consider an illustrative example for which the desired compliance matrix can be obtained both numerically and analytically. Comparison of these two solutions provides convenient benchmarks for different FEA-modeling options and also gives some practical recommendations for achieving the required accuracy.

### 4.1 Physical model

As an example, let us consider a cantilever beam of size $1000\times10\times10$ mm$^3$ with the Young's Modulus $E=2\cdot10^5\ N/mm^2$ and the Poisson's Ratio $\nu=0.266$. These data correspond to geometry and material properties of a typical robot link studied in this paper.

For this element, an analytical expression for the compliance matrix can be presented as [7]

$$k = [k_{ij}]_{6\times6} \qquad (22)$$

where non-zero elements are: $k_{35}=k_{53}=-L^2/2EI_y$, $k_{11}=L/EA$, $k_{22}=L^3/3EI_z$, $k_{33}=L^3/3EI_y$, $k_{44}=L/GJ$, $k_{55}=L/3EI_y$, $k_{66}=L/3EI_z$, $k_{26}=k_{62}=L^2/2EI_z$. Here $L$ is the length of the beam, A is its cross-section area, $I_y$, $I_z$ are the second moments, $J$ is the cross-section property.

| Mesh type in FEA model | σ, mm |
|---|---|
| Linear mesh, 2 mm | $4.59\cdot10^{-5}$ |
| Linear mesh, 1 mm | $3.87\cdot10^{-5}$ |
| Parabolic mesh, 3mm | $5.26\cdot10^{-5}$ |
| Parabolic mesh, 2 mm | $5.60\cdot10^{-5}$ |

Table 4: Parameters of the FEA-modeling noise for different mesh type

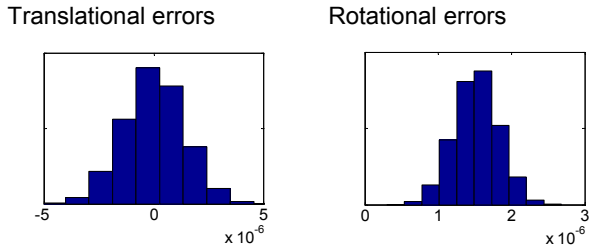

Figure 1: Histograms for the identification errors ( $a$ = 1.0 mm, b=0.1°, σ= 5×10$^{-5}$ mm)

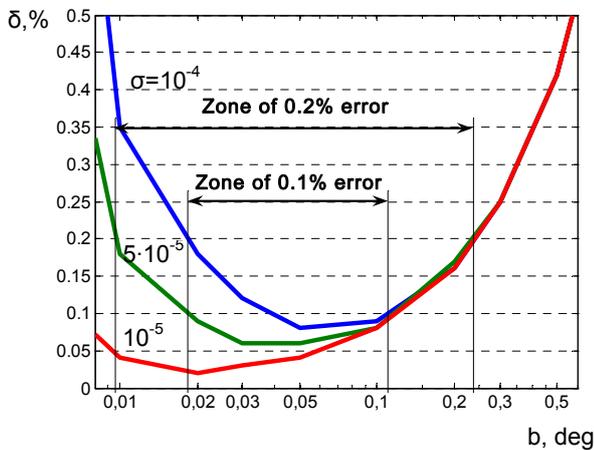

Figure 2: Identification errors for different amplitudes of the rotational deflections

### 4.2 Modeling environment

The FEA-modeling was performed using CATIA V5R16 CAD system. It is a complete tool for preparing parts geometry and generating finite element models with powerful meshing capabilities. The software was run on a computer with a 1.8 GHz processor and 1 GB memory, which impose essential constraints on the finite element dimensions even for this simple case (a single beam element) [11].

During modeling, the loads were applied at one end of the beam with the other end fully clamped. The force/torque amplitudes were determined using expression (22) and the optimal accuracy settings for the deflections 0.1…1.0 mm and 0.01…0.20° obtained in Sub-section 3.5, which yielded the following values: $F_x = 1000\,N$, $F_y = 1\,N$, $F_z = 1\,N$, $M_x = 1\,N \cdot m$, $M_y = 1\,N \cdot m$, $M_z = 1\,N \cdot m$. These loads were applied sequentially, providing 6 elementary FEA-experiments, each of which produced a single column of the compliance matrix **k**, in accordance with expression (3).

### 4.3 Meshing options

The adopted software provides two basic options for the automatic mesh generation: linear and parabolic ones. It is known that, generally, the linear meshing is faster computationally but less accurate. On the other hand, the parabolic meshing requires more computational resources while leads to more accurate results.

For the considered case study, both meshing options were examined and compared with respect to the accuracy of the obtained compliance matrix. The mesh size was gradually reduced from 5 to 1 mm, until achieving the lower limit imposed by the computer memory size. The obtained results (Table 5) clearly demonstrate advantages of the parabolic mesh, which allow achieving appropriate accuracy of 0.1% for the mesh step 2 mm using standard computing capacities. In contrast, the best result for the linear mesh is 12% and corresponds to the step of 1 mm.

### 4.4 Defining the deflection field

The developed technique operates with the deflection field corresponding to the neighborhood of the reference point (RP). As stated above, this neighborhood should be large enough to neutralize the influence of the FEA-induced errors, but its unreasonable increase may lead to violation of some essential assumptions and, consequently, to the accuracy reduction.

| Linear mesh | | | Parabolic mesh | | |
|---|---|---|---|---|---|
| 3 mm | 2 mm | 1 mm | 5 mm | 3 mm | 2 mm |
| 27% | 20% | 12% | 3.3% | 0.19% | 0.10% |

Table 5: Maximum errors in estimation of matrix k

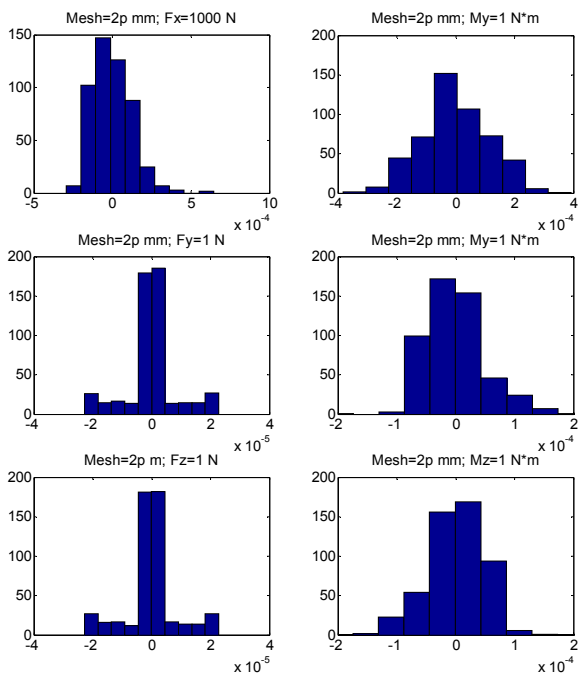

Figure 3: Residuals for stiffness model identification with parabolic mesh 2 mm

To get a realistic inference concerning this issue, a number of experiments were carried out, for different definitions of the RP-neighborhood. The obtained results show that the highest accuracy (0.1%) is achieved for the one-layer configuration of deflection field, which is composed of the nodes located on the rare edge of the examined beam. This configuration is very close to the square-type field $10 \times 10\,mm^2$ studied in sub-section 3.2. In contrast, increasing the neighborhood up to the cubic-type field $10 \times 10 \times 10\,mm^3$ leads to the identification error of about 0.08%. Hence, in practice, it reasonable to estimate the deflection values from the field corresponding to the square-type neighborhood of RP.

### 4.5 Eliminating outliers

As noticed in subsection 3.4, the FEA-modeling data may include some anomalous samples that do not obey the assumed statistical properties. This phenomena was detected in 2 of 6 experiments, (see Figure. 3) where the histograms demonstrated obvious presence of the outliers changing the regular distribution shape (local maximums around the tails). For this reason, it was applied a straightforward filtering technique that eliminated 10% the nodes corresponding to the highest residual values. This technique essentially improved the identification accuracy, the maximum error for the compliance matrix elements reduced from 0.1% to 0.05%.

It worth mentioning that here, because of the 3-dimensional nature of the problem, each node was evaluated by three residual values and was eliminated if any of the residuals was treated as an outlier. Also, the detailed analysis showed that the outliers were concentrated at the beam edges, which confirms previous assumptions concerning the FEA-induced errors.

### 4.6 Eliminating non-significant elements

According to (22), the desired compliance matrix include a number of zero elements (26 of 36), but the proposed identification procedure may produce some small non-zero values. To evaluate their statistical significance, for each element of matrix **k** it was computed the confidence interval. Relevant computations were performed using expressions for the variances of the deflections (20) and the s.t.d. value of the FEA-modeling noise, which was estimated as $\sigma = 5.6 \cdot 10^{-5}\,mm$ (by averaging for all 6 experiments). Then, the computed confidence intervals were scaled in accordance with (3), to be adopted to corresponding elements of the matrix **k**.

Using this approach, the compliance matrix was revised by assigning to zero all non-significant elements. The employed decision algorithm treated an element as non-significant if its confidence interval included zero. They allowed to detect all 26 zero elements mentioned above. It should be noted that all non-zero elements were evaluated as 'significant' ones, with essential 'safety factor' (of $10^2$ and higher).

Final results for the compliance matrix demonstrate good agreement with analytical expression (22) and confirm both accuracy of the proposed technique and its ability to detect zero-elements. Further enhancement can be achieved by the symmetrization $(\mathbf{k} + \mathbf{k}^T)/2$ that is motivated by the physical reasons.

### 4.7 Remarks and comments

Presented illustrative example that deals with a classical element (cantilever beam) confirmed validity of the developed method but also demonstrated some limitations of the FEA-modeling with respect to the stiffness analysis. In particular, it was detected some (not very essential but non negligible) non-agreement between numerical values of the applied forces/torques and their values extracted from the modeling protocol. Besides, there are a number of non-trivial issues in defining modeling options that are normally set by default. All these factors contribute to the accuracy, but practically acceptable level 0.1% can be achieved rather easily, using standard computing facilities.

## 5 STIFFNESS MODEL OF ORTHOGLIDE

Let apply the proposed methodology to the stiffness analysis of 3-d.o.f. translational mechanisms of Orthoglide-type architecture [12]. This problem was previously studied using other techniques [7], but the results were essentially different from those obtained from both the straightforward FEA-modeling and from the physical experiments.

### 5.1 Manipulator kinematics

The Orthoglide is a three d.o.f. parallel manipulator actuated by linear drives with mutually orthogonal axes. Its kinematic architecture is presented in Figure 4a and includes three identical parallel chains, which will be further referred as "legs". Each manipulator leg is formally described as PRPaR - chain, where P, R and Pa denote the prismatic, revolute, and parallelogram joints respectively. The output machinery (with a tool mounting flange) is connected to the legs in such a manner that the tool moves in the Cartesian space with fixed orientation (i.e. restricted to translational motions). The Orthoglide workspace has a regular, quasi-cubic shape. The input/output equations are simple and the velocity transmission factors are equal to one along the x, y and z direction at the isotropic configuration, like in a conventional serial PPP machine. The latter is an essential advantage for machining applications [12].

This architecture was implemented in the Orthoglide prototype, which was built in Institut de Recherche en Communications et Cybernetique de Nantes (IRCCyN) and satisfies the following design objectives: cubic Cartesian workspace of size 200×200×200 mm$^3$, Cartesian velocity and acceleration in the isotropic point 1.2 m/s and 14 m/s$^2$; payload 4 kg; transmission factor range 0.5–2.0. The legs nominal geometry is defined by the following parameters: L = 310 mm, d = 100 mm, r = 31 mm where L, d are the parallelogram length and width,

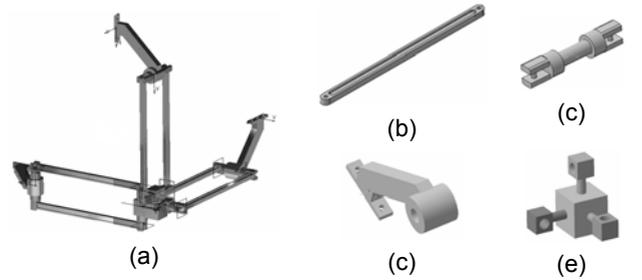

Figure 4: CAD model of Orthoglide and its principal components (a – Orthoglide, b- parallelogram bar, c – parallelogram axe d – foot, r – end-effector )

and r is the distance between the points Ci and the tool centre point P (see Figure. 4e).

The manipulator kinematics, including the direct and inverse transformations, is described in details in our previous paper [12]. Here we propose the manipulator stiffness model that, in contrast to previous works, possesses higher accuracy.

### 5.2 Stiffness of manipulator elements

The desired stiffness model for the entire manipulator (Figure 6) incorporates, as the parameters, the stiffness matrices of all principal links. Each of them was estimated using the FEA-based technique proposed in this paper. The principal components of the mechanism are presented in Figure 4, where the elements (a, b, c) are threaded as flexible ones and the element (d) is assumed to be rigid. allelogram axe d – foot, r – end-effector )

For all flexible links, the compliance matrixes were computed via the FEA-based technique proposed in this paper. Also, for comparison purposes, there were computed similar matrices corresponding to the link approximations, which are presented in Table 7  These results confirm advantages of the proposed technique, that give essential increase of accuracy. Besides, for the manipulator component (b), it was detected extreme difference (13 times) in the values of $k_{44}$ evaluated by different methods. The latter is cased by the compliancy of the joint that is taken into account in contrast to previous studies.

| Method | Compliance matrix elements | | | | | |
|---|---|---|---|---|---|---|
| | $k_{11}$ mm/N | $k_{22}$ mm/N | $k_{33}$ mm/N | $k_{44}$ rad/N·mm | $k_{55}$ rad/N·mm | $k_{66}$ rad/N·mm |
| Foot | | | | | | |
| Single-beam approximation | 3.45×10$^{-4}$ | 3.45×10$^{-4}$ | 18.1×10$^{-4}$ | 2.10×10$^{-7}$ | 2.10×10$^{-7}$ | 0.91×10$^{-7}$ |
| Four-beam approximation | 2.77×10$^{-4}$ | 4.34×10$^{-4}$ | 17.9×10$^{-4}$ | 2.11×10$^{-7}$ | 1.95×10$^{-7}$ | 0.91×10$^{-7}$ |
| FEA-based evaluation [6] (linear mesh) | 2.45×10$^{-4}$ | 3.24×10$^{-4}$ | 15.9×10$^{-4}$ | 2.07×10$^{-7}$ | 2.06×10$^{-7}$ | 1.71×10$^{-7}$ |
| FEA-based evaluation (parabolic mesh) | 2.77×10$^{-4}$ | 4.15×10$^{-4}$ | 19.4×10$^{-4}$ | 2.29×10$^{-7}$ | 2.30×10$^{-7}$ | 0.84×10$^{-7}$ |
| Parallelogram Axis | | | | | | |
| Single beam approximation | 1.34×10$^{-6}$ | 2.65×10$^{-5}$ | 2.65×10$^{-5}$ | 4.29×10$^{-8}$ | 3.18×10$^{-8}$ | 3.18×10$^{-8}$ |
| FEA-based evaluation [6] (linear mesh) | 1.99×10$^{-6}$ | 1.29×10$^{-5}$ | 1.50×10$^{-5}$ | 6.81×10$^{-6}$ | 8.23×10$^{-6}$ | 2.67×10$^{-6}$ |
| FEA-based evaluation (parabolic mesh) | 6.23×10$^{-6}$ | 2.83×10$^{-5}$ | 2.59×10$^{-5}$ | 2.77×10$^{-7}$ | 4.84×10$^{-7}$ | 1.20×10$^{-7}$ |
| Parallelogram Bar | | | | | | |
| Single-beam approximation | 3.75×10$^{-5}$ | 4.38×10$^{-2}$ | 1.09×10$^{-1}$ | 3.96×10$^{-6}$ | 3.40×10$^{-6}$ | 1.37×10$^{-6}$ |
| FEA-based evaluation [6] (linear mesh) | 4.50×10$^{-5}$ | 3.64×10$^{-2}$ | 8.01×10$^{-2}$ | 3.76×10$^{-6}$ | 2.65×10$^{-6}$ | 1.09×10$^{-6}$ |
| FEA-based evaluation (parabolic mesh) | 4.55×10$^{-5}$ | 5.08×10$^{-2}$ | 2.33×10$^{-1}$ | 2.88×10$^{-5}$ | 7.19×10$^{-6}$ | 1.50×10$^{-6}$ |

Table 7: Comparison of the link stiffness models

| Method | Compliance matrix elements | |
|---|---|---|
| | $k_{tran}$ mm/N·$10^{-4}$ | $k_{rot}$ mm/N·$10^{-7}$ |
| Lump model of Majou et al. [08] with additional passive joint | 3.68 | 2.77 |
| Lump model of Majou et al. [6] without additional passive joint | 3.68 | 1.26 |
| Overconstrained lump model with 6-dot springs [6] | 2.78 | 1.94 |
| Extended overconstrained lump model with 6-dot springs [6] | 2.93 | 2.02 |
| FEA-based evaluation [6] (linear mesh) | 3.05 | 2.05 |
| Lump model with 6-dot springs with rigid axis | 3.10 | 2.31 |
| FEA-based evaluation (parabolic mesh) | 3.02 | 2.46 |

Table 8. Comparison of the manipulator stiffness models (Orthoglide)

### 5.3 Stiffness of the manipulator

Using the obtained matrixes and applying technique from our previous paper [7], it was derived the VJM-based stiffness model of the Orthoglide manipulator. This model allows computing the stiffness matrix for any given manipulator posture, including singular configurations, without tedious re-meshing of the entire mechanism that is usually associated with conventional FEA-based methods.

Also, it was performed a straightforward FEA-based evaluation of the manipulator stiffness and compared with other modeling results (Table 8). As follows from this study, the achieved accuracy level is about 2% and essentially overcome previous approaches. Some loss of the accuracy compared to the separate links (where it is about 0.1%) is caused by neglecting some flexibility effects in the passive joints that will be in the focus of the future research.

### 6. CONCLUSIONS

The problem of accurate modeling of the manipulator stiffness arises in number of practical applications, including high-speed precision machining. Traditional analytical and semi-analytical approaches usually ignore a number of flexible effects in the manipulator mechanics while a straightforward FEA-modeling requires rather high computing resources. This paper contributes to alternative methodology that uses advantages of both analytical and numerical techniques, but requires fairly accurate stiffness matrices of the manipulator elements that are evaluated via CAD-based finite element analysis.

Proposed methodology allows essentially increase accuracy of the stiffness matrix identification by enhancing the estimation algorithms and increasing their robustness. Presented results also include detailed accuracy analysis of the stiffness identification procedures based on the statistical error model. The efficiency of the developed technique is confirmed by application examples, which deal with stiffness analysis of translational parallel manipulators and their comparison with analytical results. There are proposed practical recommendations for achieving desired accuracy of stiffness models in accordance with the requirements of particular industrial applications.

While analyzing the modeling results, there were identified several directions for prospective research activities. They include adequate modeling of the link-joint assembly and experimental verification of the stiffness models for the Orthoglide robot.

### ACKNOWLEDGEMENTS
The work presented in this paper was partially funded by the Region "Pays de la Loire", France and by the EU commission (project NEXT).